\documentclass{article}

\usepackage{microtype}
\usepackage{graphicx}
\usepackage{booktabs} 
\usepackage{url}
\usepackage{subcaption}
\usepackage[export]{adjustbox}
\usepackage{amsmath}
\usepackage{color,soul}
\usepackage{array,multirow}
\usepackage{dblfloatfix}
\usepackage{bm}
\usepackage{placeins}
\newtheorem{definition}{Definition}
\usepackage{xcolor}
\usepackage{wrapfig}
\usepackage{amssymb}
\usepackage{layouts}

\usepackage{hyperref}

\usepackage[accepted]{icml2021}

\icmltitlerunning{Influence-aware Memory}

\begin{document}

\twocolumn[
\icmltitle{Influence-aware Memory Architectures for Deep Reinforcement Learning}




\begin{icmlauthorlist}
\icmlauthor{Miguel Suau}{to}
\icmlauthor{Jinke He}{to}
\icmlauthor{Elena Congeduti}{to}
\icmlauthor{Rolf A.N. Starre}{to}
\icmlauthor{Aleksander Czechowski}{to}
\icmlauthor{Frans A. Oliehoek}{to}
\end{icmlauthorlist}

\icmlaffiliation{to}{Department of Intelligent Systems, Delft University of Technology, Delft, The Netherlands}

\icmlcorrespondingauthor{Miguel Suau}{m.suaudecastro@tudelft.nl}


\vskip 0.3in
]



\printAffiliationsAndNotice{}  


\begin{abstract}
Due to its perceptual limitations, an agent may have too little information about the state of the environment to act optimally. In such cases, it is important to keep track of the observation history to uncover hidden state. Recent deep reinforcement learning methods use recurrent neural networks (RNN) to memorize past observations. However, these models are expensive to train and have convergence difficulties, especially when dealing with high dimensional input spaces. In this paper, we propose \emph{influence-aware memory} (IAM), a theoretically inspired memory architecture that tries to alleviate the training difficulties by restricting the input of the recurrent layers to those variables that influence the hidden state information. Moreover, as opposed to standard RNNs, in which every piece of information used for estimating Q values is inevitably fed back into the network for the next prediction, our model allows information to flow without being necessarily stored in the RNN's internal memory. Results indicate that, by letting the recurrent layers focus on a small fraction of the observation variables while processing the rest of the information with a feedforward neural network, we can outperform standard recurrent architectures both in training speed and policy performance. This approach also reduces runtime and obtains better scores than methods that stack multiple observations to remove partial observability.
\end{abstract}

\section{Introduction}
\label{intro}

It is not always guaranteed that an agent will have access to a full description of the environment to solve a particular task. In fact, most real-world problems are by nature partially observable. This means that some of the variables that define the state space are hidden \cite{McCallum95PhD}. This type of problems can be modeled as \emph{partially observable Markov decision processes (POMDP)} \cite{Kaelbling96JAIR}. The model is an extension of the MDP framework \cite{Puterman94}, which, unlike the original formulation, does not not assume states to be fully observable. This implies that the Markov property is no longer satisfied. That is, future observations do not solely depend on the most recent one. 

Most POMDP methods try to extract information from the full action-observation history to disambiguate hidden state. We argue however, that in many cases, memorizing all the observed variables is costly and requires unnecessary effort. Instead, we can exploit the structure of our problem and abstract away from our history those variables that have no direct influence on the hidden ones.

Previous work on \emph{influence-based abstraction (IBA)} \cite{Witwicki10ICAPS, oliehoek2012influence} demonstrates that, in certain POMDPs, the non-Markovian dependencies in the transition and reward functions can be fully determined given a subset of variables in the history.  Hence, the combination of this subset together with the current observation forms a Markov representation that is sufficient to compute the optimal policy. In this paper, we use these theoretical insights to propose a new memory model that tries to correct certain flaws in standard RNNs that limit their effectiveness when applied to reinforcement learning (RL). We identify two key features that make our model stand apart from the most widely used recurrent architectures, LSTMs \cite{hochreiter1997long} and GRUs \cite{Cho2014Learning}:
\begin{enumerate}
    \item The input of the RNN is restricted to a subset of observation variables which in principle should contain sufficient information to estimate the hidden state. 
    \item There is a feedforward connection parallel to the recurrent layers, through which the information that is important for estimating Q values but that does not need to be memorized can flow.
\end{enumerate}

Although these two features might be overlooked as minor modifications to the standard architectures, together, they provide a theoretically sound inductive bias that brings the structure of the model into line with the problem of hidden state. Moreover, as shown in our experiments, they have an important effect on convergence, learning speed, and final performance of the agents.
\section{Related Work}
\paragraph{Partial observability:}The problem of partial observability has been extensively studied in the past. The main bulk of the work, comes from the planning community where most solutions rely on forming a belief over the states of the environment using agent's past observations \cite{ng2000pegasus,Pineau03IJCAI_PBVI,silver2010monte}. Classic RL algorithms, on the other hand, cannot directly apply the above solution due to the lack of a fully specified transition model. Instead, they learn stochastic policies that rely only on the current observation \cite{Littman94memoryless,jaakkola1995reinforcement}, or use a finite-sized history window to estimate the hidden state \cite{lin1993reinforcement, mccallum1995instance}. Curiously enough, even though the previous solutions do not scale to large and continuous state spaces, in the field of Deep RL the problem is most of the times either ignored, or naively overcome by stacking a window of past observations \cite{Mnih15Nature}. Other more sophisticated approaches incorporate external memories \cite{pmlr-v48-oh16} or use RNNs to keep track of the past history \cite{Schmidhuber91NIPS,hausknecht2015deep,jaderberg2019human}. Although this solution is much more scalable, recurrent models are computationally expensive and often have convergence difficulties when working with high dimensions \cite{Kaelbling96JAIR}. A few works, have tried to aid the RNN by using auxiliary tasks like predicting game feature information \cite{Lample17AAAI} or image reconstruction \cite{igl2018deep}. We, on the other hand, recognize that the internal structure of standard RNNs might not always be appropriate and propose a new memory architecture that is better aligned with the RL problem.
\paragraph{Attention:} One of the variants of the memory architecture we propose implements a spatial attention mechanism \cite{xu2015show} to provide the network with a layer of dynamic weights. This form of attention is different from the temporal attention mechanism that is used in seq2seq models \cite{luong2015effective, vaswani2017attention}. While the latter allows the RNN to condition on multiple past internal memories to make predictions, the spatial attention mechanism we use, is meant to filter out a fraction of the information that comes in with the observations. Attention mechanisms have recently been used in the context of Deep RL to facilitate the interpretation of the agent's behavior \cite{mott2019towards,tang2020neuroevolution} or to tackle multi-agent problems \cite{Iqbal2019Actor}. Similar to our model, the architecture proposed by Sorokin et al. \citeyearpar{sorokin2015deep} also uses an attention mechanism to find the relevant information in the game screen and feed it into the RNN. However, their model misses the feedforward connection through which the information that is useful for predicting action values but that does not need to be stored in memory can flow (see Section \ref{sec:IAMnet} for more details).
\section{Background}
The memory architecture presented in Section 4 builds on the POMDP framework, and the concept of \emph{influence-based abstraction}. For the sake of completeness, we briefly introduce each of them here and refer interested readers to \cite{Kaelbling96JAIR, oliehoek2019sufficient}.
\label{sec:IBHA}
\begin{figure*}[ht]
\hspace{2.3cm}
  \begin{subfigure}{.23\textwidth}
    \includegraphics[height=4cm,width=4cm]{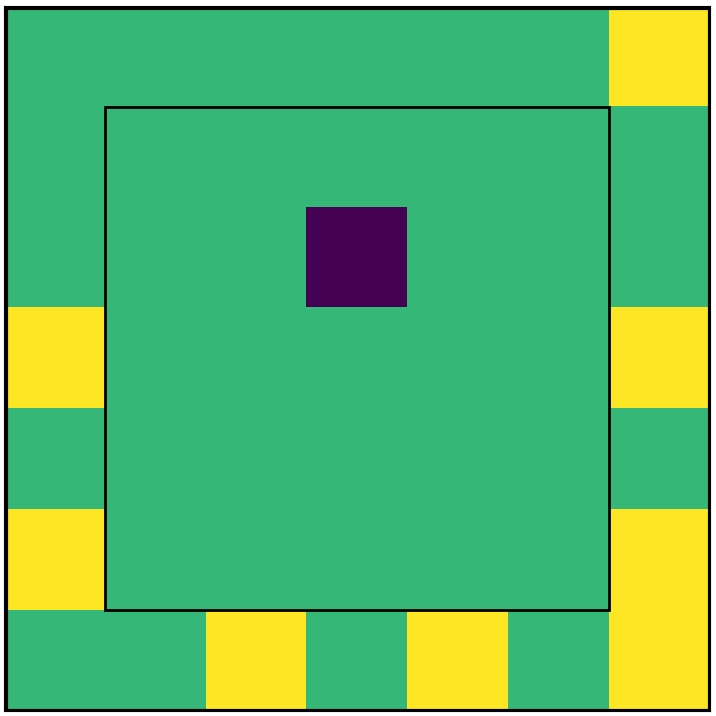}
  \end{subfigure}
  \hspace{-1cm}
  \begin{subfigure}{.67\textwidth}
    \centering
    \includegraphics[height=4cm,width=8.5cm]{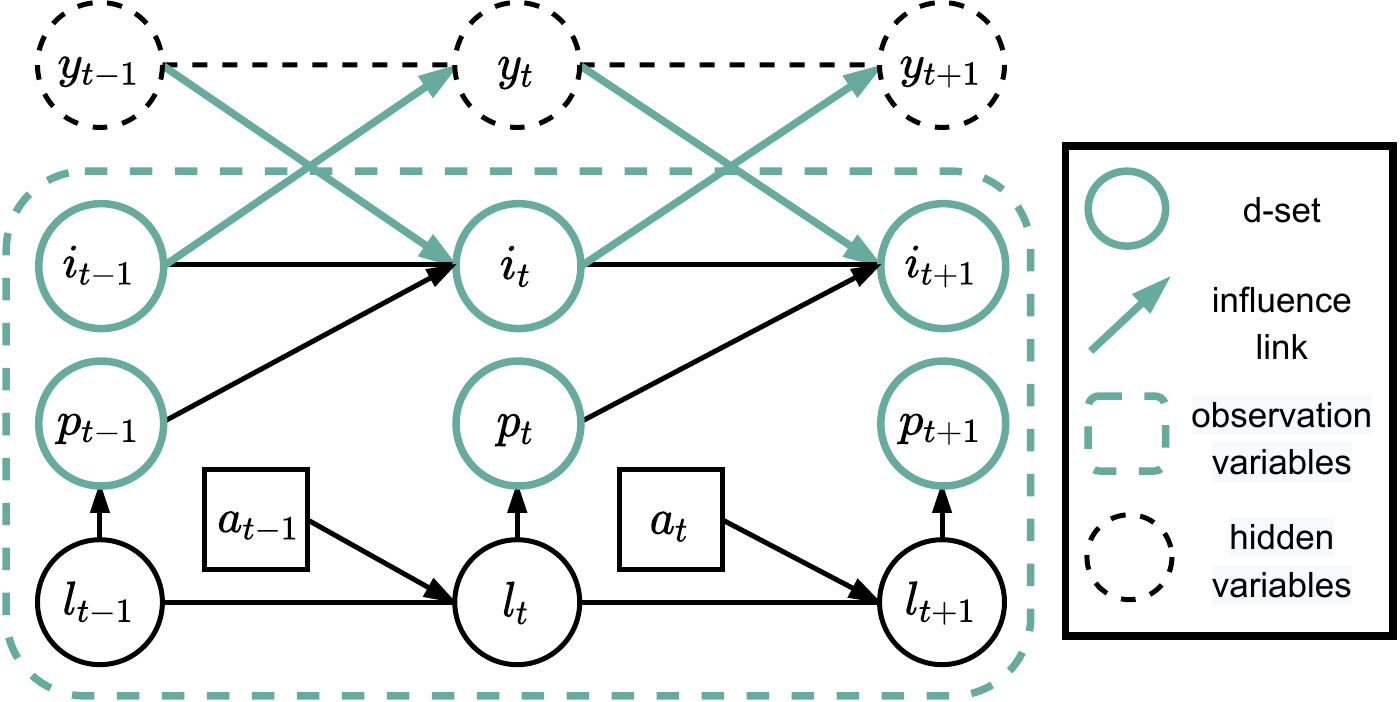}
  \end{subfigure}
\caption{Warehouse environment (left). Dynamic Bayesian Network describing the environment dynamics (right).}
\label{fig:warehouse}
\end{figure*}
\subsection{Partially Observable Markov Decision Process}
\begin{definition}[POMDP]
A POMDP is a tuple $\langle S,A,T,R,\Omega,O\rangle$ where $S$ is the state space, $A$ is the set of actions, $T$ is the transition probability function, with $T(s_t,a_t,s_{t+1})=Pr(s_{t+1}|a_t,s_t)$. $R(s_t,a_t)$ defines the reward for taking action $a_t$ in state $s_t$, $\Omega$ is the observation space, $O$ is the observation probability function, $O(a_t,s_{t+1},o_{t+1})=Pr(o_{t+1}|a_t,s_{t+1})$, the probability of observing $o_{t+1}$ after taking action $a_t$ and ending up in state $s_{t+1}$.
\end{definition}
In the POMDP setting, the task is to find the policy $\pi$ that maximizes the expected discounted sum of rewards \cite{SuttonBarto98}. Since the agent receives only a partial observation of the true state $s$, a policy that is based only on the most recent information can be arbitrarily bad \cite{Singh94ICML}. In general, the agent is required to keep track of its past experiences to make the right action choices. Policies are therefore mappings from the history of past actions and observations $h_t = \langle o_0, a_0 ..., a_{t-1}, o_t\rangle$ to actions. 

\subsection{Memory}
As mentioned in the previous section, ignoring the fact that the observations are not Markovian can lead to sub-optimal decisions. Therefore, most Deep RL methods that target partial observability use some form of memory to disambiguate hidden state. In our experiments we compare our method with the two techniques that are most widely used in practice:

\paragraph{Frame Stacking: }This simple solution was popularized by the authors of the DQN paper \cite{Mnih15Nature}, who successfully applied it to train agents on playing the Atari video games.  Although the entire game screen is provided at every iteration, some of the games, contain moving sprites whose velocity cannot be measured using only the current frame. The solution they adopted was to provide the agent with a moving window of the past 4 observations. Of course, the practicality of this approach is limited to relatively small observation spaces and short history dependencies. 

\paragraph{Recurrent Neural Networks: } A more scalable solution is to train an RNN on keeping track of the information by embedding the past action-observation history in its internal memory. However, standard recurrent neural networks, such as LSTMs \cite{hochreiter1997long} or GRUs \cite{Cho2014Learning} are known to be difficult to train and have convergence difficulties when dealing with high dimensions. The central argument of this paper is that these popular architectures, which were especially designed for a particular set of time series problems, (e.g. machine translation, speech recognition) are not the most suited for the RL task, as they fail to account for the structure many problems exhibit.
\subsection{Influence-Based Abstraction}
As mentioned in the introduction, the memory architecture we propose incorporates some of the theoretical insights developed by the framework of influence-based abstraction (IBA). Although we do not make strict use of the mathematical properties introduced below, we consider it important to include them here to motivate the memory architecture we propose. 

The fundamental idea of IBA 
is to build compact POMDP models in which hidden state variables are abstracted away by conditioning on the relevant parts of the agent's observation history. Here, rather than simplifying the transition function, we use these insights to model the agent's policy. Although according to the POMDP framework, optimal policies should condition on past actions and observations, it turns out that, in most partially observable problems, not all previous information is actually relevant. 


\paragraph{Example (Warehouse Commissioning):}
\label{sec:warehouse}
Figure 1 shows a robot (purple) which needs to fetch the items (yellow) that appear with probability $0.05$ on the shelves at the edges of the $7\times7$ grid representing a warehouse. The robot receives a reward of $+1$ every time it collects an item. The added difficulty of this task is that item orders get canceled if they are not collected before 8 timesteps since they appear. Thus, the robot needs to maintain a time counter for each item and decide which one is best to go for.

The dynamics of the problem are represented by the dynamic Bayesian network (DBN) \cite{Pearl88, boutilier1999decision} in Figure \ref{fig:warehouse} (right), where $l_t$ denotes the robot's current location in the warehouse, and $i_t$ and $p_t$ are binary variables indicating if the item order is active and whether or not the robot is at the item pick-up location. The hidden variable $y_t$ is the item's time counter\footnote{For simplicity, we only include a single item in the DBN in Figure \ref{fig:warehouse}. The dynamics of all the other of the items in the warehouse are analogous.}, to which the robot has no access. The robot can only infer the time counter based on past observations. To do so, however, it does not need to remember the full history, but only whether or not a given item order was active at a particular timestep. More formally, inspecting the DBN, we see that $y_{t+1}$ is only indirectly influenced by the agent's past location $l_{t-1}$ via $p_{t-1}$ and the item variable $i_{t}$. Therefore, we say that $y_{t+1}$ is conditionally independent of $l_{t-1}$ given $p_{t-1}$ and $i_t$,
\begin{equation}
    Pr(y_{t+1}|l_{t-1}, p_{t-1}, i_t) = Pr(y_{t+1}|p_{t-1}, i_t)
\end{equation}
This means that in order to infer the hidden variable $y$ at any timestep it is sufficient to condition on the past values of $p$ and $i$. The history of these two variables, highlighted in green in Figure \ref{fig:warehouse}, constitutes the d-separating set (d-set).

\begin{definition}[D-separating set] The d-separating set is a subset of variables $d_t$ from the agent's action-observation history $h_t$, such that the hidden variables $y_t$ and the remaining parts of the history $h_t \setminus d_t$ are conditionally independent given $d_t$: $\Pr(y_t|h_t) = P(y_t | d_t, h_t\setminus d_t)  = Pr(y_t | d_t).$ This conditional independence can be tested using the notion of d-separation \cite{Bishop06book}.
\label{def:d-set}
\end{definition}
\begin{figure*}[ht]
\vskip -0.1in
\begin{center}
\centerline{\includegraphics[width=.8\textwidth]{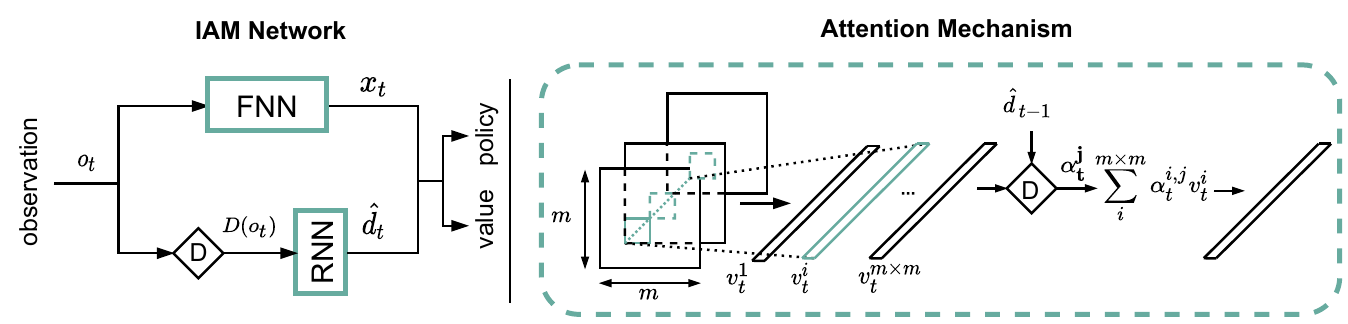}}
\caption{Influence-aware Memory network architecture (left). Diagram of the attention mechanism for image data (right)}
\label{fig:IAMnet}
\end{center}
\vskip -0.3in
\end{figure*}


\label{sec:background}
\section{Influence-aware Memory}
The properties outlined in the previous section, are not we unique to the warehouse example. In fact, as we show in our experiments, it is often the case in partially observable problems that only a fraction of the observation variables influence the hidden state. This does not necessarily imply that the agent can completely ignore the rest of the information. In the warehouse example, the robot's current location, despite being irrelevant for inferring the hidden state, is in fact crucial for estimating the current action values.

The Bellman equation for the optimal action value function $Q^*$ of a POMDP can be expressed in terms of the history of actions and observations $h_t$ as
\begin{equation}
\small
\begin{split}
   Q^*(h_t, a_t) &= R(h_t,a_t) \\
   & + \sum_{o_{t+1}} \Pr(o_{t+1}|h_t,a_t) \max_{a_{t+1}} Q^*(h_{t+1},a_{t+1}),
\end{split}
   \label{eq:POMDP}
\end{equation}
where $R(h_t, a_t) = \sum_{s_t} \Pr(s_t|h_t)R(s_t,a_t)$ is the expected immediate reward at time $t$ over the set of possible states $s_t$ given a particular history $h_t$.

According to IBA, we can replace the dependence on the full history of actions and observations $h_t$ by a dependence on the d-set $d_t$ (Definition \ref{def:d-set}),
\begin{equation}
\small
\begin{split}
   Q^*(\langle d_t, o_t \rangle, a_t) &=  R(\langle d_t, o_t \rangle,a_t) \\
   + \sum_{o_{t+1}}  \Pr(o_{t+1}|&\langle d_t, o_t \rangle,a_t) \max_{a_{t+1}} Q^*(\langle d_{t+1}, o_{t+1} \rangle, a_{t+1}),
   \label{eq:IBA}
\end{split}
\end{equation}
and $d_{t+1} \triangleq \langle d_t, D(o_{t+1})\rangle$, where $D(\cdot)$ is the d-set selection operator, which chooses the variables in $o_{t+1}$ that are added to $d_{t+1}$. Note that, although $d_t$ contains enough information to estimate the hidden state (Equation \ref{eq:IBA} and Definition \ref{def:d-set}), $o_t$ is still needed to estimate $Q$. Hence, given the tuple $\langle d_t, o_t \rangle$ we can write
\begin{equation}
\small
    Q^*(h_t, a_t) = Q^*(\langle d_t, o_t \rangle, a_t),
    \label{eq:QhQd}
\end{equation}
The upshot is that in most POMDPs the combination of $d_t$ and $o_t$ forms a Markov representation that the agent can use to find the optimal policy. Unfortunately, in the RL setting, we are normally not provided a fully specified DBN to determine the exact d-set. Nonetheless, in many problems like in our warehouse example it is not difficult to make an educated guess about the variables containing sufficient information to predict the hidden ones. The network architecture we present in the next section enables us to select beforehand what variables the agent should memorize. This is however not an prerequisite since, as we explain in Section \ref{sec:approxdsets}, we can also force the RNN to find such variables by restricting its capacity.
\subsection{Influence-aware Memory Network}\label{sec:IAMnet}
The Influence-aware Memory (IAM) architecture we propose is depicted in Figure \ref{fig:IAMnet}. The network tries to encode the ideas of IBA as inductive biases with the goal of being able to learn policies and value functions more effectively. Following from \eqref{eq:QhQd}, our architecture implements two separate networks in parallel: an FNN, which processes the entire observation, 
\begin{equation}
    x_t = F_{\text{fnn}}(o_t),
\end{equation} 
and an RNN, which receives only $D(o_{t+1})$ and updates its internal state, 
\begin{equation}
    \hat{d}_{t} = F_{\text{rnn}}(\hat{d}_{t-1},D(o_{t+1})),
\end{equation}
where we use the notation $\hat{d}_t$ to indicate that the d-set is embedded in the RNN's internal memory. The output of the FNN $x_t$ is then concatenated with $\hat{d}_t$ and passed through two separate linear layers which compute values $Q(\langle x_t,\hat{d}_t \rangle, a_t)$ and action probabilities $\pi(\langle x_t,\hat{d}_t \rangle, a_t)$.

\begin{figure}[ht]
\vskip -0.1in
\begin{center}
\centerline{\includegraphics[width=1.07\columnwidth]{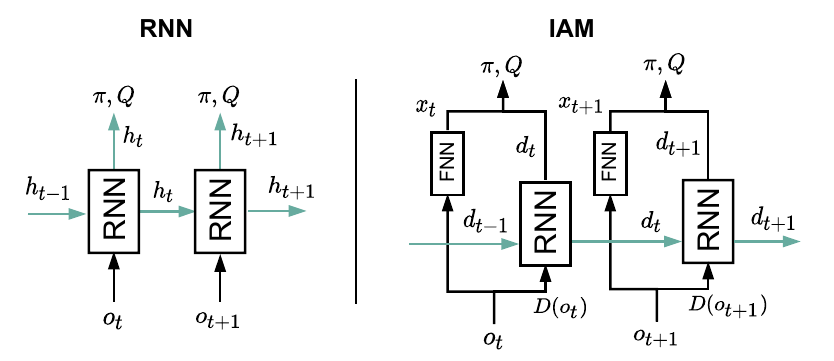}}
\caption{Information flow in standard RNNs (left) compared to IAM (right). The diagram on the left shows that the same vector $h_t$ that is used for estimating $\pi$ and $Q$ is also part of the input for the next prediction (green arrows). On the other hand, in the IAM architecture there is another vector $x_t$ coming out from the FNN, which is only used for estimating $\pi$ and $Q$ at time $t$ and is not stored in memory. Hence, the RNN in IAM is free to include in $\hat{d}_t$ only the information that the agent needs to remember.}
\label{fig:RNNvsIAM}
\end{center}
\vskip -0.3in
\end{figure}

\paragraph{IAM vs. standard RNNs:} We try to facilitate the task of the RNN by feeding only the information that, in principle, should be enough to uncover hidden state. This is only possible thanks to the parallel FNN channel, which serves as an extra gate through which the information that is useful for predicting action values but that does not need to be stored in memory can flow. This is in contrast to the standard recurrent architectures that are normally used in Deep RL (e.g. LSTM, GRU, etc.), which suffer from the fact that every piece of information that is used for estimating values is inevitably fed back into the network for the next prediction. Intuitively, standard RNNs face a conflict: they need to choose between ignoring those variables that are unnecessary for future predictions, risking worse $Q$ estimates, or processing them at the expense of corrupting their internal memory with irrelevant details. Figure \ref{fig:RNNvsIAM} illustrates this idea by comparing the information flow in both architectures.

Finally, since the recurrent layers in IAM are freed from the burden of having to remember irrelevant information, they can be dimensioned according to the memory needs of the problem at hand. This translates into networks that combine regular size FNNs together with small RNNs.

\paragraph{Image data:}If our agent receives images rather than feature vectors, we first preprocess the raw observations $o$ with a CNN, $F_{\text{cnn}}(o_t) = \mathbf{v_t}$ and obtain $m \times m$ vectors $v$ of size $N$, where $N$ is the number of filters in the last convolutional layer and $m \times m$ the dimensions of the 2D output array of each filter (Figure \ref{fig:IAMnet} right).  Fortunately, since the convolution operator preserves the topology of the input, each of these vectors corresponds to a particular region of the input image. Thus, we can still use domain knowledge to to choose which vectors should go into the RNN.

\subsection{Learning approximate d-sets }
\label{sec:approxdsets}
Having the FNN channel can help detach the RNN from the task of estimating the current $Q$ values. However, without the d-set selection operator $D$, nothing prevents the information that does not need to be remembered from going through the RNN. Although, as we show in our first two experiments, it is often possible for the designer to guess what variables directly influence the hidden state information, it might not always be so straightforward. In such cases, rather than manually selecting the d-set, the agent will have to learn $D$ from experience. In particular, we add a linear layer before the RNN, to act as information 
bottleneck \cite{tishby2015deep} and filter out all that information that is irrelevant:
\begin{equation}
    \hat{D}_{A}(o_t) = A o_t
    \label{eq:approx-dset}
\end{equation}
where $\hat{D}$ indicates that the operator is learned rather than handcrafted and $A$ is a matrix of weights of size $K \times N$, where $N$ is the number of observation variables (the number of filters in the last convolutional layer when using images) and $K$ is a hyperparameter that determines the dimensions of the output.  
The matrix A needs to be computed differently depending on the nature of the problem:
\paragraph{Static d-sets:} If the variables that must go into the d-set do not change from one timestep to another. That is, if $D$ always needs to choose the same subset of observation variables, as occurs in the warehouse example, we just need a fixed matrix $A$ to filter all observations in the same way. $A$ can be implemented as a separate linear layer before the RNN or we can just directly restrict the size the RNN's input layer.

\paragraph{Dynamic d-sets:} If, on the other hand, the variables that must go into the d-set do change from one timestep to another, we use a multi-head spatial attention mechanism \cite{xu2015show, vaswani2017attention} to recompute the weights in every iteration. Thus we write $A_t$ to indicate that the weights can now adapt to $o_t$ and $\hat{d}_{t-1}$. The need for such dynamism can be easily understood by considering the Atari game of breakout. To be able to predict where the ball will be next, the agent does not need to memorize the whole set of pixels in the game screen, but only the ones containing the ball, whose location differs in every observation and hence the need of a varying matrix $A_t$. Specifically, for each row $i$ in $A_t$, each element $\alpha^{i,j}_t$ is computed by a two-layer fully connected network that takes as input the corresponding element in the observation vector $o_i$ and $\hat{d}_{t-1}$, followed by a softmax operator. Figure \ref{fig:IAMnet} is a diagram of how each of the attention heads operates for the case of using as input the output of the CNN $\mathbf{v_t}$ instead of the observation vector $o_t$. Please refer to the Appendix for more details about the technical implementation of this mechanism.

Note that the above solutions would not be able to filter out the information that is only useful for the current $Q$ estimates without the parallel FNN connection (Figure \ref{fig:RNNvsIAM}). We would also like to stress that these mechanisms are by no means guaranteed to find the optimal d-set. Nonetheless, as shown in our experiments, they constitute an effective inductive bias that facilitates the learning process.



\label{sec:IAM}
\section{Experiments}
We empirically evaluate the performance of our memory architecture on the warehouse example (Section \ref{sec:IBHA}), a traffic control task, and the \emph{flickering} version of the Atari video games \cite{hausknecht2015deep}. The goal of our experiments is:
\begin{enumerate}
    \item \textbf{Learning performance and convergence: } Evaluate whether our model improves over standard recurrent architectures. We compare learning performance, convergence and training time.
    \item \textbf{High dimensional observation spaces:} Show that our solution scales to high dimensional problems with continuous observation spaces.
    \item \textbf{Learning approximate d-sets: } Demonstrate the advantages of restricting the input to the RNN and compare the relative performance of learning vs. manually specifying the d-sets.
    \item \textbf{Architecture analysis: } Analyze the impact of the architecture on the learned representations by inspecting the network hidden activations.
\end{enumerate}

\subsection{Environments}
Below is a brief description of the three domains on which we evaluate our model. Please refer to the Appendix for more details.

\begin{figure}[ht]
\begin{center}
\centerline{\includegraphics[width=0.5\columnwidth]{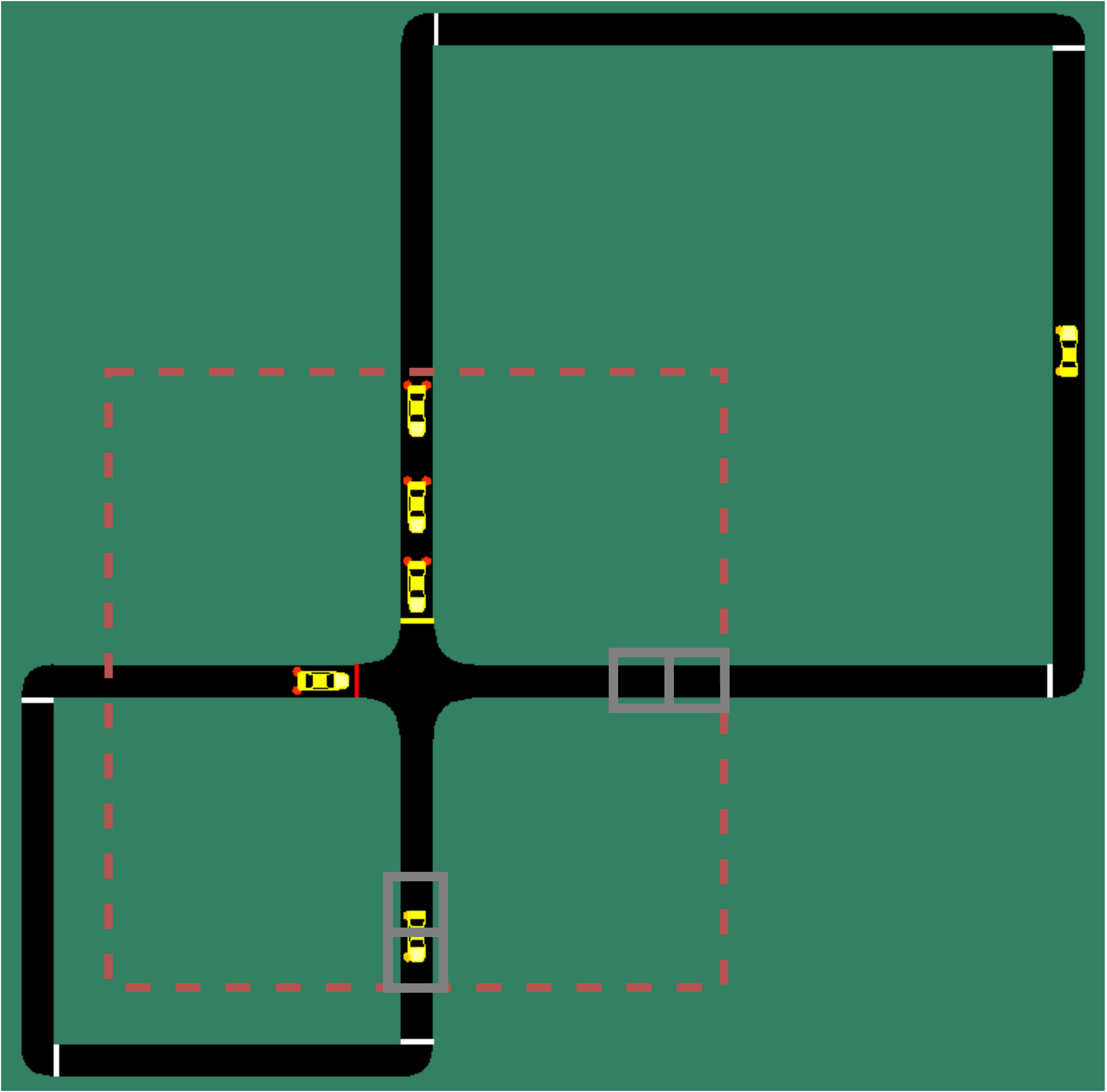}}
\caption{Traffic control environment. Cars are only visible when they enter the red box. The small grey boxes are the variables that we feed into the RNN.}
\label{fig:traffic_control}
\end{center}
\vskip -0.4in
\end{figure}

\paragraph{Warehouse:} This is the same task we describe in our example in Section \ref{sec:background}. The observations are a combination of the agent's location (one-hot encoded vector) and the 24 item binary variables. In the experiments where d-sets are manually selected, the RNN in IAM only receives the latter variables while the FNN processes the entire vector.
\paragraph{Traffic Control: }
In this environment \cite{SUMO2018}, the agent must optimize the traffic flow at the intersection in Figure \ref{fig:traffic_control}. The agent can take two different actions: either switching the traffic light on the top to green, which automatically turns the other to red, or vice versa. The observations are binary vectors that encode whether not there is a car at a particular location. Cars are only visible when they enter the red box. There is a 6 seconds delay between the moment an action is taken and the time the lights actually switch. During this period the green light turns yellow, and no cars are allowed to cross the road. Agents need to anticipate cars entering the red box and switch the lights in time for them to continue without stopping. This forces the recurrent models to remember the location and the time at which cars left the intersection and limits the performance of agents with no memory\footnote{Videos showing the results of the traffic control experiment can be found at \url{https://tinyurl.com/wc3jpf4}}. 
In the experiments where d-sets are manually selected, the RNN in IAM receives the last two elements in each of the two vectors encoding the road segments (i.e. 4 bits in total). The location of these elements is indicated by the small grey boxes in Figure \ref{fig:traffic_control}. This information should be sufficient to infer hidden state. 
\paragraph{Flickering Atari: } In this version of the Atari video games \cite{bellemare13arcade} the observations are replaced by black frames with probability $p=0.5$. This adds uncertainty to the environment and makes it more difficult for the agent to keep track of moving elements. The modification was introduced by Hausknecht \& Stone \citeyearpar{hausknecht2015deep} to test their recurrent version of DQN and has become the standard benchmark for Deep RL in POMDPs \cite{Zhu17arxiv, igl2018deep}.
\subsection{Experimental Setup}
We compare IAM against two other network configurations: A model with no internal memory that uses frame stacking (FNN) and a standard recurrent architecture (LSTM). All three models are trained using PPO \cite{schulman2017proximal}. For a fair comparison, and in order to ensure that both types of memory have access to the same amount of information, the sequence length parameter in the recurrent models (i.e. number of time steps the network is unrolled when updating the model) is chosen to be equal to the number of frames that are fed into the FNN baseline. We evaluate the performance of our agents at different time steps during training by calculating the mean episodic return. The results are averaged over three random seeds. A table containing the full list of hyperparameters used for each domain and for each of the three architectures, together with a detailed description of the tuning process is provided in the Appendix.

\subsection{Learning performance and convergence}
\begin{figure}[ht]
\vskip -0.3in
\begin{center}
\centerline{\includegraphics[width=1.2\columnwidth]{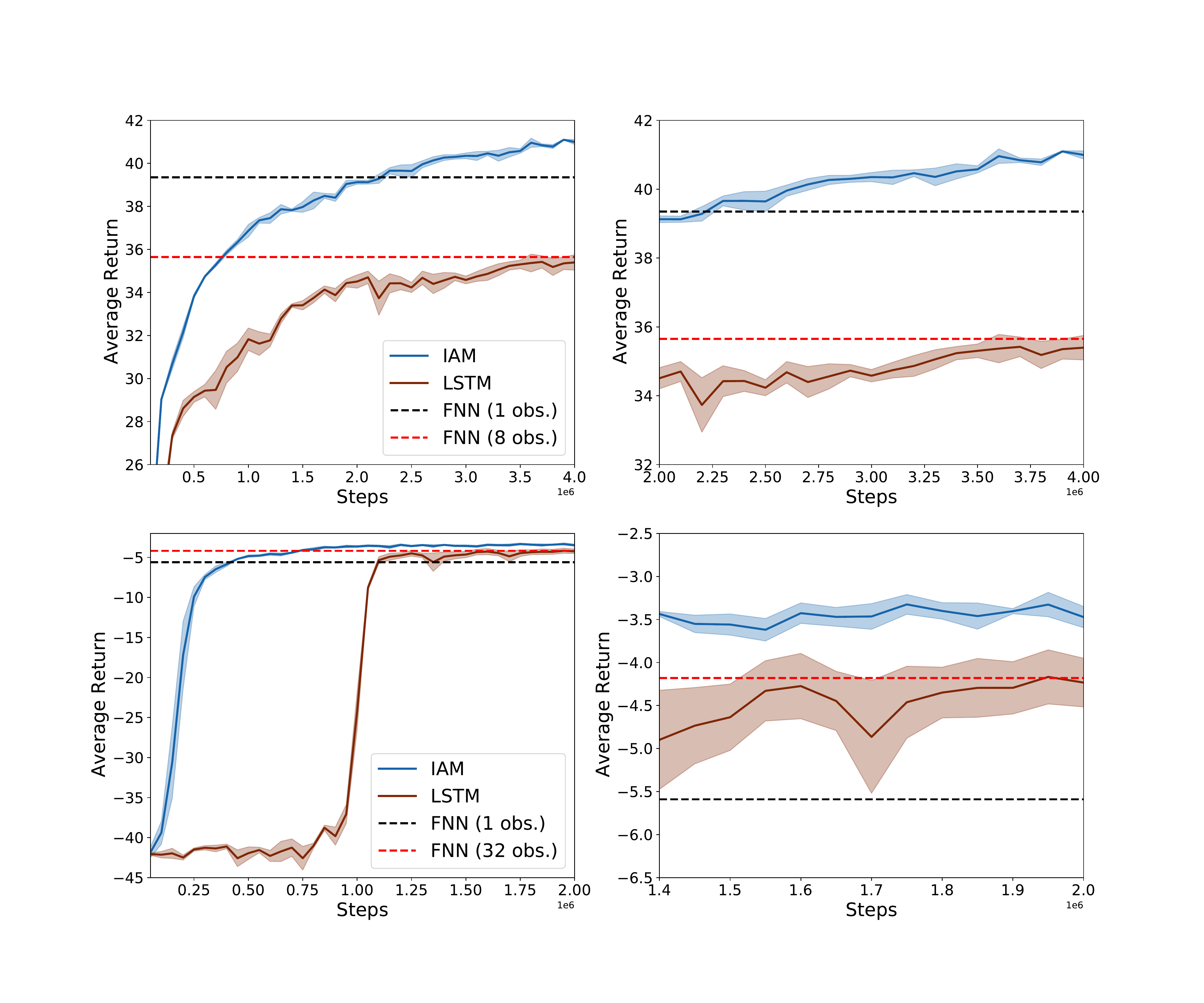}}
\vskip -0.3in
\caption{Average return and standard deviation during training of IAM and LSTM as a function of the number of timesteps on the warehouse (top) and the traffic (bottom) environments. For ease of visualization, the plots on the right are zoomed in versions of the ones on the left. The dashed horizontal lines indicate the final performance of FNNs with (red) and without memory (black).}
\label{fig:learning_curves}
\end{center}
\vskip -0.45in
\end{figure}
We first evaluate the performance of our model on the warehouse and traffic control environments. Although the observation sizes are relatively small compared to most deep RL benchmarks ($73$ and $30$ variables respectively), the two tasks are quite demanding memory-wise. In the warehouse environment, the agent is required to remember for how long each of the items has been active. In the traffic domain, cars take 32 timesteps to reappear again in the red box when driving around the big loop (Figure \ref{fig:traffic_control}).

Figure \ref{fig:learning_curves}, shows the learning curves of IAM and LSTM in the two environments. While both LSTM and IAM reach similar levels of performance on the traffic control task the LSTM network takes much longer to converge (bottom). On the other hand, in the warehouse environment, IAM clearly outperforms the LSTM baseline (top). The final scores obtained by FNNs with (red) and without memory\footnote{Please note that, although the optimal policy in these two environments requires memory, memoryless policies can still reach a decent performance level.} (black) (i.e. observation stacking) are also included for reference. These results are strong evidence that the parallel feedforward channel in IAM is indeed helping overcome the convergence difficulties of LSTMs, by bypassing the recurrent layers (Section \ref{sec:IAMnet}).

\begin{figure}[ht]
\vskip -0.3in
\begin{center}
\centerline{\includegraphics[width=1.2\columnwidth]{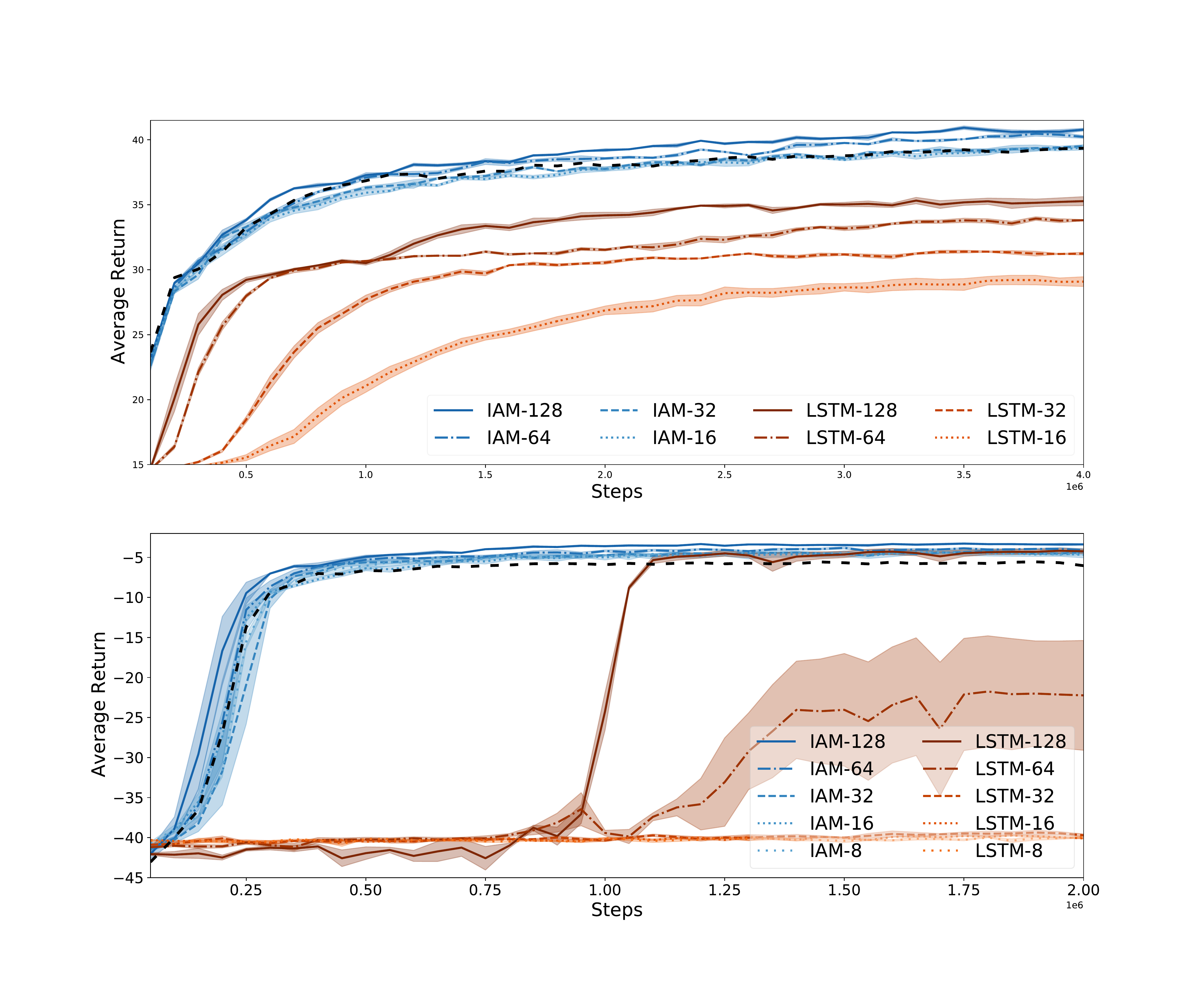}}
\vskip -0.3in
\caption{Average return and standard deviation during training of IAM (learned $\hat{D}$) and LSTM for various recurrent layer sizes on the warehouse (top) and the traffic (bottom) environments. The dashed black lines are the learning curves of FNNs without memory.}
\label{fig:IAMvsLSTM}
\end{center}
\vskip -0.3in
\end{figure}

Figure \ref{fig:IAMvsLSTM} is a performance comparison of LSTM and IAM for various recurrent layer sizes\footnote{For a fair comparison, the reported results for IAM correspond to networks where $\hat{D}$ is learned and not manually specified. The labels indicate the total number of recurrent neurons. Both LSTMs and IAMs have also feedfoward layers of equivalent size. A detailed description of how these were chosen so as not to interfere with the results is provided in the Appendix.}. While the best recurrent layer size for the LSTM baseline is $\bm{128}$ in both domains, the size of the recurrent model of IAM can be brought down to $\bm{64}$ for the warehouse environment and just $\bm{8}$ recurrent neurons for the traffic task, and still outperform the memoryless FNN baseline (dashed black curve). This, of course, translates into a significant reduction in the total number of weights and computational speedups.
A full summary of the average runtime for each architecture, along with a description of the computing infrastructure used is given in the Appendix.
\subsection{High dimensional observation spaces}
The advantage of IAM over LSTMs and FNNs becomes even more apparent as the dimensionality of the problem increases. Table \ref{tab:atari_scores} compares the average scores obtained in Flickering Atari by the FNN and LSTM baselines with those of IAM. Both IAM and LSTM receive only $1$ frame. The sequence length parameter is set to $8$ time steps for the two networks. The FNN model, on the other hand, receives the last $8$ frames as input. The learning curves are shown in the Appendix together with the results obtained in the original games and the average runtime.
\begin{table}
\vskip -0.1in
\centering
  \caption{Average final score on the Flickering Atari games for each of the three network architectures and standard deviation. Bold numbers indicate the best results on each environment.}
\vspace{2mm}
\resizebox{\columnwidth}{!}{
\centering
  \begin{tabular}{c|c|c|c}
& FNN (8 frames) & LSTM & IAM \\\cline{1-4}
Breakout & $26.57 \pm  1.51$  & $21.32 \pm 0.45$ &$\mathbf{83.10 \pm 5.29}$  \\\hline
Pong & $18.07 \pm 0.06$ & $-20.25 \pm 0.03$ &  $\mathbf{20.07 \pm 0.11}$ \\\hline
Space Invaders & $\mathbf{854.93 \pm 11.64}$
&$520.44 \pm 9.41$ & $\mathbf{834.66 \pm 21.23}$ \\\hline
Asteroids & $1393.75\pm 11.28$
&$1424.87 \pm 5.23$ & $\mathbf{2281.63 \pm 63.92}$ \\\hline
MsPacman & $\mathbf{2388.03 \pm 167.03}$
&$1081.11 \pm 293.79$ & $\mathbf{2326.04 \pm 31.53}$ \\\hline
\end{tabular}
}
\vskip -0.2in
\label{tab:atari_scores}
\end{table}
\begin{figure*}
\centering
    \begin{subfigure}[b]{0.2\textwidth}
    \includegraphics[width=\textwidth]{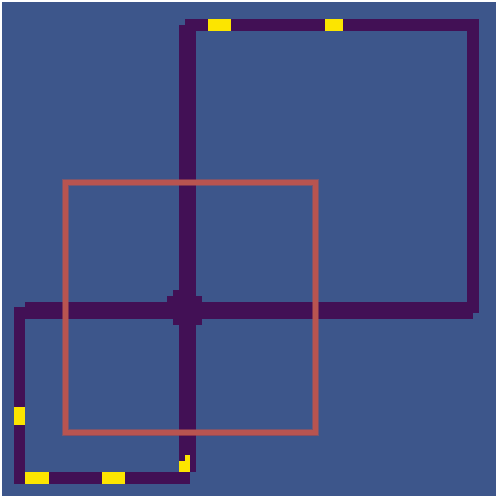}
  \end{subfigure}
  \begin{subfigure}[b]{0.2\textwidth}
    \includegraphics[width=\textwidth]{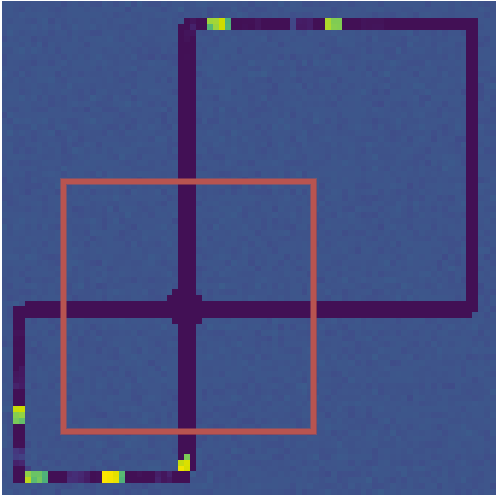}
  \end{subfigure}
  \begin{subfigure}[b]{0.59\textwidth}
    \includegraphics[width=\textwidth]{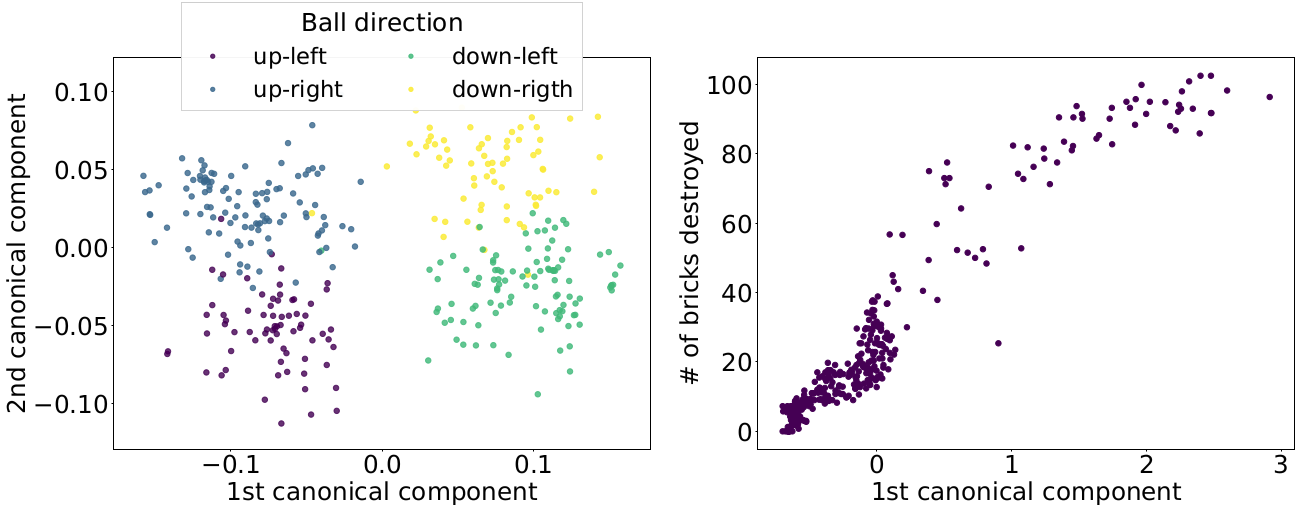}
  \end{subfigure}
\caption{Example of a full game screen and the reconstruction made by the memory decoder (left). RNN's internal memories $\hat{d}$ projected onto the two first canonical components, colors indicate the direction of the velocity vector (second from the right). FNN's outputs $x$ projected onto the first canonical component against the number of bricks destroyed (rightmost).}
\label{fig:CCA}
\vspace{-3mm}
\end{figure*}
\subsection{Learning approximate d-sets.}
\begin{table}
\vskip -0.3in
\centering
  \caption{Comparison between manually selecting the d-set and learning it, and between using static and dynamic weights.}
 \vspace{2mm}
\resizebox{\columnwidth}{!}{
\centering
  \begin{tabular}{c|c|c|c}
& Manual & Learned (static) & Learned (dynamic) \\\cline{1-4}
Warehouse &  $\bm{40.99 \pm 0.20}$ & $40.76 \pm 0.12$ & - \\\hline
Traffic Control &  $-3.76 \pm 0.12$ & $\bm{-3.37 \pm 0.09}$ & - \\\hline
Breakout &  -  & $26.17 \pm 0.82$  &$\bm{83.10 \pm 5.29}$  \\\hline
Pong & - & $\bm{19.72 \pm 0.27}$ &  $\bm{20.07 \pm 0.11}$ \\\hline
Space Invaders & - & $583.12 \pm 8.76$ & $\bm{834.66 \pm 21.23}$ \\\hline
Asteroids & - & $1841.87 \pm 26.09$ & $\bm{2281.63 \pm 63.92}$ \\\hline
MsPacman & - & $2097.97 \pm 34.71$ & $\bm{2326.04 \pm 31.53}$ \\\hline
\end{tabular}
}
\vskip -0.2in
\label{tab:approx_dsets}
\end{table}
As explained in Section \ref{sec:approxdsets}, if the optimal d-set is static, like in the warehouse and traffic environments, we might be able to learn $\hat{D}$ by simply restricting the size of the RNN.
The first two rows in Table \ref{tab:approx_dsets}, show a performance comparison between manually selecting our d-set, and forcing the network weights to filter out the irrelevant variables by limiting its capacity. The problem needs to be treated with a bit more care in cases where the variables that influence the hidden state change from one observation to another, as occurs in the Atari games. In such situations, just restricting the size of the RNN is not sufficient since the weights are static, and hence unable to settle for any particular subset of pixels in the game screen (Section \ref{sec:approxdsets}). The last two columns in Table \ref{tab:approx_dsets} are the scores obtained in Flickering Atari with static weights (2nd column) and with the dynamic weights computed by the attention mechanism (3rd column). Manual selection is not feasible in the Atari domain.
\subsection{Architecture Analysis}\label{sec:architectureanalysis}
\paragraph{Decoding the agent's internal memory:}\label{sec:decoder}
We evaluated if the information stored in the agent's internal memory after selecting the d-set and discarding the rest of the observation variables was sufficient to uncover hidden state. To do so, we trained a decoder on predicting the full game screen given the encoded observation $x_t$ and $\hat{d}_t$, using a dataset of images and hidden activations collected after training the policy. The image on the leftmost of Figure \ref{fig:CCA} shows an example of the full game screen, from which the agent only receives the region delimited by the red box. The second image from the right shows the prediction made by the decoder. Note that although everything outside the red box is invisible to the agent, the decoder is able to make a fair reconstruction of the entire game screen based on the d-set encoded in the agent's internal memory $\hat{d}$. This implies that IAM can capture the necessary information and remember how many cars left the intersection and when without being explicitly trained to do so\footnote{\label{note1}A video of this experiment where we use the decoder to reconstruct an entire episode can be found at \url{https://tinyurl.com/y9cvuz7l}. More examples are provided in the Appendix.}. 
\paragraph{Analysis of the hidden activations:}\label{sec:abstractinfluence}
Finally, we  used Canonical Correlation Analysis (CCA) \cite{hotelling1992relations} to measure the correlation between the network hidden activations when playing Breakout and two important game features: ball velocity and number of bricks destroyed. 
The projections of the hidden activations onto the space spanned by the canonical variates are depicted in the two plots on the right of Figure \ref{fig:CCA}. The scatter plot on the left shows four distinct clusters of hidden memories $\hat{d}_t$. Each of these clusters corresponds directly to one of the four possible directions of the velocity vector. The plot on the right, shows a clear uptrend. High values of the first canonical component of $x_t$ correspond to frames with many missing bricks. While the FNN is taking care of the information that does not need to be memorized (i.e. number of bricks destroyed) the RNN is focused on inferring hidden state variables (i.e. ball velocity). 
More details about this experiment are given in the Appendix.

\label{sec:experiments}
\section{Conclusion}
The primary goal of this paper was to reconcile neural network design choices with the problem of partial observability. We studied the underlying properties of POMDPs and developed a new memory architecture that tries to decouple hidden state inference from value estimation. Influence-aware memory (IAM) connects an FNN and an RNN in parallel. This simple solution allows the RNN to focus on remembering just the essential pieces of information. This is not the case in other recurrent architectures. Gradients in LSTMs and GRUs need to reach a compromise between two, often competing, goals. On the one hand, they need to provide good $Q$ estimates and on the other, they should remove from the internal memory everything that is irrelevant for future predictions. Our model enables the designer to select beforehand what variables the agent should memorize. This is however not an prerequisite since, as shown in our experiments, we can force the RNN to find such variables by restricting its capacity. We also investigated a solution for those problems in which the variables influencing the hidden state information differ from one observation to another.
Our results suggest that while standard architectures have severe convergence difficulties, IAM can even outperform methods that stack multiple frames to remove partial observability. Finally, aside from the clear benefits in learning performance, our analysis of the network hidden activations suggests that the inductive bias introduced in our memory architecture enables the agent to choose what to remember.


\bibliography{bibliography}
\bibliographystyle{icml2021}

\end{document}